\documentclass[runningheads]{llncs}
\usepackage[T1]{fontenc}
\usepackage{graphicx}
\usepackage{amsmath,amsfonts}
\usepackage{algorithmic}
\usepackage{algorithm}
\usepackage{array}
\usepackage{color}
\usepackage{marvosym}

\def\ft#1{\textbf{#1}}
\def\st#1{\underline{#1}}

\begin{document}
\title{Toward Better SSIM Loss for Unsupervised Monocular Depth Estimation}
\author{Yijun Cao\orcidID{0000-0002-1934-2698} \and
Fuya Luo\orcidID{0000-0001-9027-8468} \and
Yongjie Li\textsuperscript{(\Letter)}\orcidID{0000-0002-7395-3131}}

\authorrunning{Y. Cao et al.}
%
\institute{
The MOE Key Laboratory for Neuroinformation, the School of Life Science and Technology, 
University of Electronic Science and Technology of China, Chengdu 610054, China.\\
\email{\{yijuncaoo, luofuya1993\}@gmail.com} \\
\email{liyj@uestc.edu.cn}}

%
\maketitle              
%
\begin{abstract}
    Unsupervised monocular depth learning generally relies on 
    the photometric relation among temporally adjacent images. 
    Most of previous works use both mean absolute error (MAE)
    and structure similarity index measure (SSIM) with conventional form as training loss.
    However, they ignore the effect of different components in the SSIM function 
    and the corresponding hyperparameters on the training.
    To address these issues, this work proposes a new form of SSIM.
    Compared with original SSIM function, 
    the proposed new form uses addition rather than multiplication to combine 
    the luminance, contrast, and structural similarity related components in SSIM. 
    The loss function constructed with this scheme helps result in smoother gradients 
    and achieve higher performance on unsupervised depth estimation.
    We conduct extensive experiments to determine the relatively optimal combination of parameters for our new SSIM.
    Based on the popular MonoDepth approach, the optimized SSIM loss function can remarkably outperform the baseline 
    on the KITTI-2015 outdoor dataset.

\end{abstract}

\section{Introduction}
Single Image Depth Estimation (SIDE) is a critical task in the field of computer vision.
It contributes to many other tasks, 
e.g., edge detection \cite{cao-tmm21}, scene reconstruction \cite{zhu2022nice}, 
object detection \cite{peng2021saliency} and visual odometry \cite{cao2023learning}.
Recent learning-based SIDE can be generally divided into 
supervised \cite{liuva,Eigen-nips14} and unsupervised approaches \cite{Godard-iccv19,cao2023learning}.
Supervised SIDE requires ground-truth depth map as training labels, 
whose process would be costly and time-consuming. 
Usupervised methods learn depth maps using photometric consistency from consecutive monocular or stereo image pairs.
Recently, though unsupervised SIDE has seen great progresses, 
there still exists a large performance gap comparied with supervised methods. 
We argue that this gap comes from the following four aspects: 
occlusion and illumination inconsistency among adjacent images, inaccurate pose estimation (especially for the dynamic regions), 
loss function and network architecture.

In the past few years, researchers mainly focus on addressing the first two issues, i.e., reducing the effect of
inconsistency among adjacent images \cite{Godard-iccv19,Luo-pami20} and pursuing accurate pose \cite{Zhao-cvpr20,cao2023learning},
while ignoring the last two issues.
Thus, based on popular pixel-wise mean absolute error (MAE) and structure similarity index measure (SSIM),
this work analyzes the effect of different coefficients on SSIM and designs a more suitable SSIM loss function for unsupervised SIDE.
In addition, we found that the sub-pixel convolution \cite{shi2016real} is better than conventional interpolation schemes
in the processing of depth upsampling.
A large number of experiments on KITTI dataset \cite{Geiger-ijrr13} demonstrates that 
the proposed loss function and network are better than baseline and many existing unsupervised SIDE methods.

\section{Related Works}
\subsection{Supervised Monocular Depth Learning}
With the development of convolutional Neural Networks (CNNs), a variety of models have been proposed to learn monocular depth 
in a supervised manner \cite{Eigen-iccv15,liuva}. 
These approaches usually take 
a single image as input and use RGB-D camera or LIDAR as ground truth labels. 
Supervised methods were usually to design a better network for capturing structural information in monocular images,
by using ranking or ordinal relation constraint \cite{zoran2015learning}, surface normal constraint \cite{hu2019revisiting}
or other heuristic refinement constraint \cite{yuan2022neural}.
However, the supervised methods require labeled ground truth, which are 
expensive to obtain in natural environments.

\subsection{Unsupervised Monocular Depth Learning}
 More recent works have begun to approach the SIDE task in a unsupervised way. 
A pioneering work is SfMLearner \cite{Zhou-cvpr17}, which learns depth and ego-motion jointly by minimizing photometric loss in an unsupervised manner.
This pipeline has inspired a large amount of follow-up works. To deal with moving objects breaking the assumption of static scenes, 
many works \cite{Yin-cvpr18,Luo-pami20} employ the consistency of forward-backward optical \cite{Zhao-cvpr20}, depth-optical \cite{Zou-eccv18}, 
or depth-depth \cite{Bian-nips2019} flow to mask dynamic objects.
Several methods developed new frameworks by changing training strategies and adding supplementary constraints \cite{Godard-iccv19}
and collaborative competition \cite{Ranjan-cvpr19}.
More recently, several researchers \cite{Zhao-cvpr20,Yang-cvpr20} have tried to combine a geometric algorithm into the deep learning architecture, and obtained 
better depth and VO estimations by training with only two frames in a video sequence. 
Compared with previous works, the proposed method improves the loss function 
and proposes to use sub-pixel convolution \cite{shi2016real} to replace nearest interpolation as upsampling approach
aiming to obtain accurate and smooth SIDE.

\section{Method} 
\subsection{Unsupervised depth learning pipeline}
The aim of learning based monocular depth estimation is to predict a pixel-aligned depth map $D_t$ with an input image $I_t$
via a network with the parameter set $\theta$, 
\begin{equation}
    D_t=DepthNet(I_t; \theta).
\end{equation}

For computing re-projection error, or named photometric error, 
we need to calculate the ego-motion from image $I_t$ to $I_{t+1}$,
\begin{equation}
    T_{t\rightarrow t+1}=PoseNet(I_t, I_{t+1}; \theta),
\end{equation}
where $T_{t \rightarrow t'}\in \mathcal{S} \mathcal{E} (3)$ is the ego-motion 
in three dimensional special euclidean group from time $t$ to $t+1$.

Given the depth $D_t$ and ego-motion $T_{t\rightarrow t+1}$, 
for a pixel $p_t$ in $I_t$, the corresponding pixel $p_{t+1}$ in $I_{t+1}$ can be found
through camera perspective projection, which are consistent for static scenes. 
Formally, the relationships can be written as
\begin{equation}
    p_{t+1} = KT_{t \rightarrow t+1}D_{t}(p_{t})K^{-1}p_{t}, \\
\end{equation}
where $K$ and $D_{t}(p_{t})$ denote the camera intrinsic and the depth in $p_{t}$, respectively.

After computing the corresponding $p_t$ and $p_{t+1}$, the synthetic image $I'_{t \leftarrow t+1}$ can be warped using $I_{t+1}$.
Similarly, we can use $I_{t-1}$ to synthesize $I'_{t \leftarrow t-1}$ with pose  $T_{t\rightarrow t-1}$.
Then, unsupervised training of depth is realized by minimizing the photometric error 
between the raw and the synthetic images, like Monodepth2 \cite{Godard-iccv19}:
\begin{equation}
   \mathcal{L}_{p} = \frac{1}{N}\sum min(r(I_t, I'_{t \leftarrow t+1}), r(I_t, I'_{t \leftarrow t-1})),
   \label{eq:old_pe}
\end{equation}
where $N$ is the total number of pixels in the image $I_t$.
The function $r(\cdot, \cdot)$ is the metric between the target and synthetic images.

Pixel-level color matching alone is unstable and ambiguous. Therefore, an
edge-aware smoothness term is often applied for regularization \cite{Luo-pami20}:
\begin{equation}
   \mathcal{L}_{s} = \frac{1}{N} \sum |\frac{\partial d_t}{\partial x}| e^{-|\frac{\partial I_t}{\partial x}|} +
   |\frac{\partial d_t}{\partial y}| e^{-|\frac{\partial I_t}{\partial y}|},
   \label{eq:smoothness} 
\end{equation}
where $d_t = 1/D_t$ can be considered as disparity.

\subsection{Improved Photometric Error}
The main loss used for training depth prediction model in unsupervised manner is photometric error (Equation \ref{eq:old_pe}).
Most previous works \cite{Luo-pami20,Godard-iccv19} usually define the pixel-wise image metric $r(I, I')$
using the combination of SSIM \cite{Zhou-tip04} with MAE,
\begin{equation}
   r(I, I') = \frac{\kappa}{2}(1-SSIM(I, I')) + (1-\kappa)|I-I'|,
   \label{eq:l1_ssim}
\end{equation}
where $\kappa$ is the weight and usually set to $0.85$ \cite{Luo-pami20,Godard-iccv19}. 
SSIM is the function of structure similarity index measure \cite{Zhou-tip04} for evaluating the similarity between two images,
which is consists of three key features of the differences between two images: 
luminance $L$, contrast $C$ and structure $S$,
\begin{equation}
    SSIM = L^\alpha \cdot C^\beta \cdot S^\gamma,
\end{equation}
where symbols SSIM, L, C and S are functions that we omit its independent variable ($I$ and $ I'$) for a more concise description.
$\alpha,\beta,\gamma$ represent the proportion of different characteristics in the SSIM measure.
The value of SSIM is in the range of $[-1, 1]$, the values closer to $1$ mean the higher similarities between $I$ and $I'$,
so to make the SSIM available for the gradient descent algorithm, we need to make a simple transformation $\frac{1}{2}(1-SSIM)$.

\begin{figure}[t]
    \centering
    \includegraphics[width=12cm]{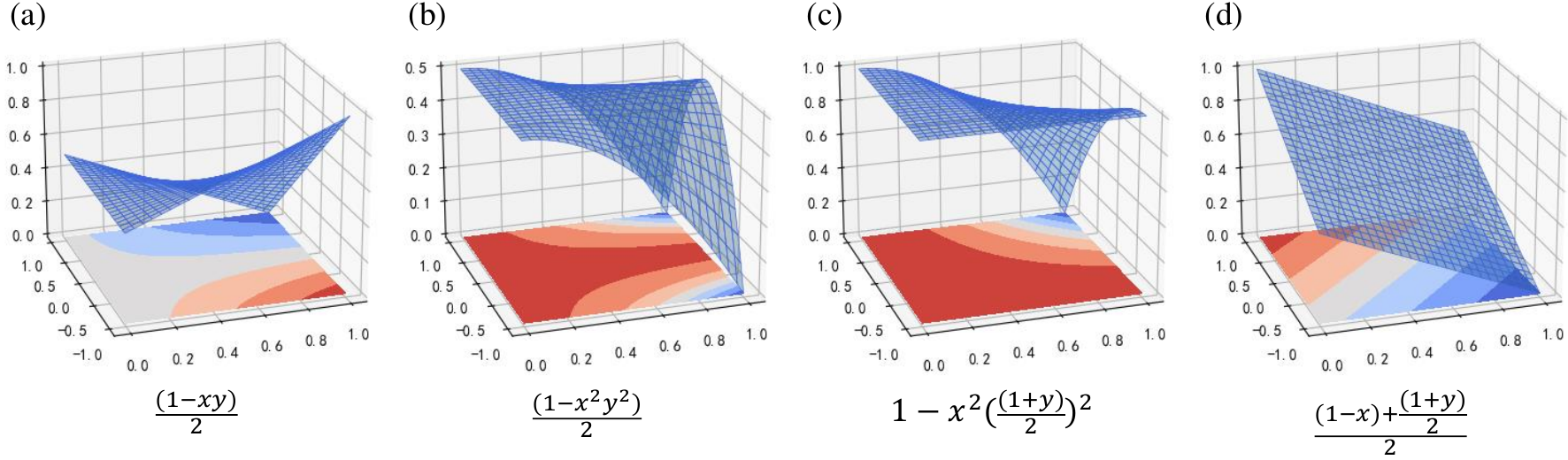}
    \caption{
    Four toy examples to illustrate the effect of different components of the SSIM on the output.
    $x$ indicates luminance and contrast parts, whose outputs are in the range $[0,1]$.
    $y$ indicates structure part, whose output is in the range $[-1,1]$.
    }
    \label{fig:examples}
\end{figure}

To measure the similarity of two images, we usually compute the metrics by sliding window across the image and then average them.
Typically, most previous works has used $\alpha=1,\beta=1,\gamma=1$ and set window size to 3.
However, the regular settings are not always the best settings. 
In our experiments (see Table \ref{table:SSIM_m}), 
a significant performance improvement can be obtained by simply adjusting these three weights $\alpha$, $\beta$ and $\gamma$.
Note that luminance $L$, contrast $C$ and structure components $S$ are in the range $[0, 1]$, $[0, 1]$ and $[-1, 1]$, respectively.
If $\gamma=1$ is set to even, then its minimum value is not unique (Figure \ref{fig:examples}(b))
and we need to give structure component a transformation before calculating the exponent,
\begin{equation}
    SSIM_m = 1 - L^\alpha \cdot C^\beta \cdot (\frac{1}{2}(1+S))^\gamma.
\end{equation}
Figure \ref{fig:examples}(a-c) show three toy examples of the SSIM outputs changing with exponent weights.
Compared with Figure \ref{fig:examples}(a) and \ref{fig:examples}(c), we can see as the exponential weights get larger, 
their outputs become steeper near zero and more flatten out away from zero.
Table \ref{fig:examples} demonstrates that larger exponential weights result in better performances on the KITTI dataset.

The original form of SSIM is the combination of components ($L,C,S$) by multiplication 
and the weight of each component is weighted by an exponent.
It may be not a good choose as a loss function to train a deep learning model.
Figure \ref{fig:examples}(a-c) show that combining by multiplication leads to the uneven gradient distribution.
In some initializations, there may be some parameters that do not converge well 
due to the small gradients backpropagated based on the SSIM outputs.
Therefore, we propose to combine the different components using addition,
\begin{equation}
    SSIM_a = w_l(1-L) + w_c(1-C) + w_s(1-\frac{1}{2}(1+S)),
\end{equation}
where $w_l$, $w_c$ and $w_s$ are the weights of components $L$, $C$ and $S$, respectively.
Figure \ref{fig:examples}(d) shows the example of the function $SSIM_a$, 
we can see that the gradient of the result is equal everywhere by a combination with addition rather than multiplication.

\begin{figure}[ht]
    \centering
    \includegraphics[width=12cm]{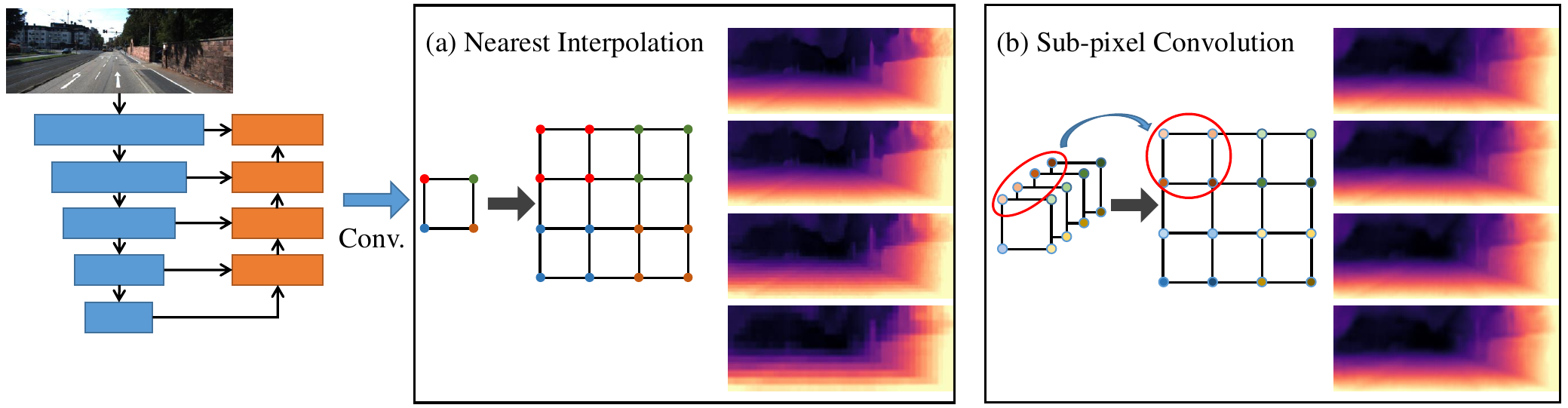}
    \caption{
    The network architecture. (a) the nearest upsampling module used in the baseline network..
    (b) the sub-pixel convolution upsampling module used in our network.
    }
    \label{fig:network}
 \end{figure}

\subsection{Multi-scale Depth Estimation}
The depth network is similar to the architecture in \cite{Godard-iccv19}, which adopts 
encoder-decoder design with skip connections and five-scale side outputs.
The encoder is ResNet18 \cite{He-cvpr16} without full connection layers;
at each scale, the decoder consists of 2 convolutional layers with the kernel size of 3 and 
1 convolutional layer with the kernel size of 1.
Different from the previous network \cite{Godard-iccv19},
we use sub-pixel convolution \cite{shi2016real} rather than nearest interpolation to upsample the low resolution depth map to higher.

Sub-pixel convolution \cite{shi2016real} is the strategy converting information from channel to spatial.
For example, given the original low-resolution depth map of size $H/2 \times W/2 \times r^2$, 
the channels of each pixel are rearranged into an $r \times r$ spatial region,
so that the depth map of size $H/2 \times W/2 \times r^2$ is rearranged into a high-resolution of $H \times W \times 1$.
Figure \ref{fig:network} shows the difference between the nearest interpolation and sub-pixel convolution method.
Note that although the technique of sub-pixel convolution is already used on task of super-resolution, 
we introduce it to solve the unsupervised SIDE task and achieve better results compared with many previous works \cite{Godard-iccv19,Luo-pami20}.

\section{Experiments}
\textbf{Dataset.} The KITTI dataset \cite{Geiger-ijrr13} commonly used for the task of depth estimation provides videos of 200
street scenes captured by RGB cameras, with sparse depth ground truths captured by laser scanner. 
For depth, training was done on the KITTI raw \cite{Bian-nips2019,Godard-iccv19}
and the frames were resized to $640\times192$ pixels. The depth was evaluated on the Eigen's testing split \cite{Eigen-nips14}.

\textbf{Implementation details.} The Adam \cite{Kingma-iclr15} optimizer was used, the learning rate was set to $10^{-4}$, and the
batch size to 8. The training epochs was set to 20.
We used warmup strategy to update learning rate, the warmup step was set to 1000.
For a more fair comparision, we set the random seeds to 1234 on 
\textit{PyTorch} \cite{paszke2019pytorch} and \textit{NumPy} \cite{harris2020array} package.

\textbf{Metrics.} We adopted the standard metrics (Abs Rel, Sq Rel, RMSE, RMSE log, 
$\delta_1<1.25$, $\delta_2<1.25^2$, $\delta_3<1.25^3$) to evaluate the depth from 0-80 meters. 
Detailed definitions can be found in \cite{Luo-pami20}.

\begin{table}[ht]
    \centering
    \caption{Ablation studies on $SSIM_m$. All the methods were trained in a self-supervised manner with monocular data. 
    Bolded numbers are the best metrics.}
    \label{table:SSIM_m}
    \begin{tabular}{p{0.8cm}p{0.8cm}p{0.8cm}ccccccccc}
       \hline
       \multicolumn{3}{c}{}& \multicolumn{4}{c}{Error}& \multicolumn{3}{c}{Accuracy} \\
       $\alpha$ & $\beta$ & $\gamma$ & AbsRel$\downarrow$& SqRel$\downarrow$& RMS$\downarrow$& RMSlog$\downarrow$& $<1.25\uparrow$& $<1.25^2\uparrow$& $<1.25^3\uparrow$\\
       \hline
        \multicolumn{3}{l}{Monodepth2 \cite{Godard-iccv19}} 
                    & 0.115      & 0.903 & 4.863 & 0.193 & 0.877 & 0.959 & 0.981 \\

        \multicolumn{3}{l}{Baseline}  & 0.118      & 0.868      & 4.856      & 0.194 &      0.869 &      0.957 &      0.981 \\
        \hline
        2 & 1 & 1 & 0.115      & 0.860      & 4.895      & 0.193 &      0.871 &      0.957 &      0.981 \\
        3 & 1 & 1 & 0.116      & 0.861      & 4.880      & 0.193 &      0.872 &      0.958 &      0.981 \\
        \hline
        1 & 2 & 1 & 0.115      & \ft{0.835} & \ft{4.823} & 0.193 &      0.871 &      0.958 & \ft{0.982} \\
        1 & 3 & 1 & 0.116      & 0.856      & 4.894 &      0.194 &      0.871 &      0.957 &      0.981 \\
        \hline
        1 & 1 & 2 & \ft{0.114} & 0.836      & 4.862 &      0.193 &      0.873 &      0.958 &      0.981 \\
        1 & 1 & 3 & 0.115      & 0.868      & 4.830 & \ft{0.192} & \ft{0.874} & \ft{0.959} &      0.981 \\
        \hline
        2 & 2 & 2 & 0.115      & 0.838      & 4.834 & \ft{0.192} & \ft{0.874} &      0.958 &      0.981 \\
        3 & 3 & 3 & 0.115      & 0.829      & 4.826 & \ft{0.192} &      0.871 &      0.958 &      0.981 \\
       \hline

    \end{tabular}
\end{table}

\begin{table}[ht]
    \centering
    \caption{Ablation studies on $SSIM_a$. All the methods were trained in a self-supervised manner with monocular data. 
    Bolded numbers are the best metrics.}
    \label{table:SSIM_a}
    \begin{tabular}{p{0.6cm}p{0.6cm}p{0.6cm}p{0.6cm}ccccccccc}
       \hline
       \multicolumn{4}{c}{}& \multicolumn{4}{c}{Error}& \multicolumn{3}{c}{Accuracy} \\
       $w_1$ & $w_l$ & $w_c$ & $w_s$ & AbsRel$\downarrow$& SqRel$\downarrow$& RMS$\downarrow$& RMSlog$\downarrow$& $<1.25\uparrow$& $<1.25^2\uparrow$& $<1.25^3\uparrow$\\
       \hline
        \multicolumn{4}{l}{Monodepth2 \cite{Godard-iccv19}} 
                    & 0.115      & 0.903 & 4.863 & 0.193 & 0.877 & 0.959 & 0.981 \\
        \multicolumn{4}{l}{Baseline}  & 0.118      & 0.868      & 4.856      & 0.194 &      0.869 &      0.957 &      0.981 \\
        \hline
        0.3 & 0.4 & 0.4 & 0.5 & 0.116      & 0.836      & 4.856      & 0.194      & 0.871      & 0.958      & 0.981 \\
        0.3 & 0.4 & 0.4 & 0.6 & 0.115      & 0.842      & 4.879      & 0.193      & 0.871      & 0.958      & \ft{0.982} \\
        0.3 & 0.4 & 0.4 & 0.7 & 0.116      & 0.834      & 4.833      & 0.193      & 0.871      & 0.958      & 0.981 \\
        \hline
        0.3 & 0.4 & 0.5 & 0.5 & 0.116      & 0.850      & 4.871      & 0.194      & 0.870      & 0.958      & 0.981 \\
        0.3 & 0.4 & 0.5 & 0.6 & 0.115      & 0.828      & 4.859      & 0.193      & 0.871      & 0.958      & 0.981 \\
        0.3 & 0.4 & 0.5 & 0.7 & 0.115      & 0.832      & 4.828      & 0.192      & \ft{0.873} & \ft{0.959} & 0.981 \\
        \hline
        0.3 & 0.5 & 0.5 & 0.5 & 0.115      & \ft{0.816} & 4.846      & 0.194      & 0.871      & 0.958      & 0.981 \\
        0.3 & 0.5 & 0.5 & 0.6 & \ft{0.114} & \ft{0.816} & 4.823      & \ft{0.191} & \ft{0.873} & \ft{0.959} & \ft{0.982} \\
        0.3 & 0.5 & 0.5 & 0.7 & 0.115      & 0.849      & 4.858      & 0.193      & 0.872      & 0.958      & 0.981 \\
        \hline
        0.3 & 0.6 & 0.5 & 0.5 & \ft{0.114} & 0.819      & 4.834      & 0.192      & 0.872      & 0.958      & \ft{0.982} \\
        0.3 & 0.6 & 0.5 & 0.6 & 0.115      & 0.831      & 4.854      & 0.193      & 0.872      & 0.958      & 0.981 \\
        0.3 & 0.6 & 0.6 & 0.5 & 0.117      & 0.837      & 4.898      & 0.195      & 0.867      & 0.957      & 0.981 \\
        0.3 & 0.6 & 0.6 & 0.6 & 0.116      & 0.830      & 4.854      & 0.194      & 0.870      & 0.958      & 0.981 \\
        \hline 
        0.4 & 0.4 & 0.4 & 0.6 & 0.116      & 0.865      & 4.886      & 0.194      & 0.872      & 0.957      & 0.981 \\
        0.4 & 0.4 & 0.5 & 0.6 & 0.116      & 0.842      & 4.840      & 0.193      & 0.872      & 0.958      & 0.981 \\
        0.4 & 0.4 & 0.5 & 0.7 & \ft{0.114} & 0.826      & \ft{4.822} & 0.192      & 0.872      & \ft{0.959} & 0.981 \\
        \hline
        0.4 & 0.5 & 0.5 & 0.5 & 0.115      & 0.836      & 4.853      & 0.194      & 0.871      & 0.957      & 0.981 \\
        0.4 & 0.5 & 0.5 & 0.6 & 0.115      & 0.817      & 4.846      & 0.193      & 0.871      & 0.957      & 0.981 \\
        0.4 & 0.5 & 0.5 & 0.7 & 0.115      & 0.823      & 4.828      & 0.193      & \ft{0.873} & 0.958      & \ft{0.982} \\
        \hline
        0.4 & 0.6 & 0.5 & 0.5 & 0.116      & 0.819      & 4.836      & 0.192      & \ft{0.873} & 0.958      & 0.981 \\
        0.4 & 0.5 & 0.6 & 0.5 & 0.117      & 0.843      & 4.856      & 0.194      & 0.870      & 0.957      & 0.981 \\
       \hline
    \end{tabular}
\end{table}

\begin{table}[ht]
    \centering
    \caption{Ablation studies on upsampling methods.}
    \label{table:subpixel}
    \begin{tabular}{p{0.6cm}p{0.6cm}p{0.6cm}p{0.6cm}ccccccccc}
       \hline
       \multicolumn{4}{c}{}& \multicolumn{4}{c}{Error}& \multicolumn{3}{c}{Accuracy} \\
       $w_1$ & $w_l$ & $w_c$ & $w_s$ & AbsRel$\downarrow$& SqRel$\downarrow$& RMS$\downarrow$& RMSlog$\downarrow$& $<1.25\uparrow$& $<1.25^2\uparrow$& $<1.25^3\uparrow$\\
       \hline
        0.3 & 0.5 & 0.5 & 0.6                  & \ft{0.114} & 0.816      & 4.823      & 0.191      & \ft{0.873} & \ft{0.959} & 0.982 \\
        \multicolumn{4}{l}{w/ sub-pixel conv.} & 0.116      & \ft{0.776} & \ft{4.811} & \ft{0.190} & 0.862      & 0.957      & \ft{0.983} \\
        \hline
        0.4 & 0.4 & 0.5 & 0.7                  & \ft{0.114} & 0.826      & 4.822      & 0.192      & \ft{0.872} & \ft{0.959} & 0.981 \\
        \multicolumn{4}{l}{w/ sub-pixel conv.} & 0.116      & \ft{0.784} & \ft{4.793} & \ft{0.189} & 0.863      & 0.957      & \ft{0.983} \\
       \hline
       0.4 & 0.5 & 0.5 & 0.7                   & \ft{0.115} & 0.823      & 4.828      & 0.193      & \ft{0.873} & \ft{0.958} & 0.982 \\
        \multicolumn{4}{l}{w/ sub-pixel conv.} & \ft{0.115} & \ft{0.770} & \ft{4.813} & \ft{0.188} & 0.863      & 0.957      & \ft{0.983} \\
       \hline
    \end{tabular}
\end{table}

\begin{table}[ht]
    \centering
    \caption{Comparing multi-scale performance of nearest upsample and sub-pixel convolution.
    We use combining $SSIM_a$ and MAE as training loss with $w_1=0.4$, $w_l=0.5$, $w_c=0.5$, $w_s=0.7$.}
    \label{table:resolution}
    \begin{tabular}{ccccccccccc}
       \hline
       \multicolumn{2}{c}{}& \multicolumn{4}{c}{Error}& \multicolumn{3}{c}{Accuracy} \\
       Resolution & Bilinear & AbsRel$\downarrow$& SqRel$\downarrow$& RMS$\downarrow$& RMSlog$\downarrow$& $<1.25\uparrow$& $<1.25^2\uparrow$& $<1.25^3\uparrow$\\
       \hline
        $1/2$ & \checkmark & \ft{0.115} & 0.823      & 4.828      & 0.193      & \ft{0.873} & \ft{0.958} & 0.982 \\
        $1/2$ &            & \ft{0.115} & \ft{0.770} & \ft{4.813} & \ft{0.188} & 0.863      & 0.957      & \ft{0.983} \\
       \hline
       $1/4$ & \checkmark  & \ft{0.115} & 0.818      & 4.833      & 0.192      & \ft{0.870} & \ft{0.958} & 0.982 \\
        $1/4$ &            & \ft{0.115} & \ft{0.770} & \ft{4.812} & \ft{0.188} & 0.863      & 0.957      & \ft{0.983} \\
       \hline
       $1/8$ & \checkmark  & 0.119      & 0.833      & 4.925      & 0.195      & 0.862      & 0.956      & 0.982 \\
        $1/8$ &            & \ft{0.115} & \ft{0.771} & \ft{4.820} & \ft{0.188} & \ft{0.863} & \ft{0.957} & \ft{0.983} \\
       \hline
       $1/16$ & \checkmark & 0.130      & 0.943      & 5.203      & 0.203      & 0.844      & 0.952      & 0.981 \\
        $1/16$ &           & \ft{0.116} & \ft{0.780} & \ft{4.844} & \ft{0.189} & \ft{0.862} & \ft{0.957} & \ft{0.983} \\
       \hline
    \end{tabular}
\end{table}

\begin{table}[ht]
    \centering
    \caption{Quantitative comparison of the proposed method with existing methods.
    SSC denotes the $SSIM_a$ and sub-pixel convolution.
    Bolded and underlined numbers are respectively the best and second-best metrics.}
    \label{table:sota}
    \begin{tabular}{lccccccc}
       \hline
       \multicolumn{1}{c}{}& \multicolumn{4}{c}{Error}& \multicolumn{3}{c}{Accuracy} \\
       Methods & AbsRel$\downarrow$& SqRel$\downarrow$& RMS$\downarrow$& RMSlog$\downarrow$& $<1.25\uparrow$& $<1.25^2\uparrow$& $<1.25^3\uparrow$\\
       \hline
        SfMLearner \cite{Zhou-cvpr17}           & 0.183&      1.595&      6.709&      0.270&      0.734&      0.902&      0.959 \\
        CC \cite{Ranjan-cvpr19}                 & 0.140&      1.070&      5.326&      0.217&      0.826&      0.941&      0.975 \\
        EPC++ \cite{Luo-pami20}                 & 0.141&      1.029&      5.350&      0.216&      0.816&      0.941&      0.976 \\
        SC-SfMLearner \cite{Bian-nips2019}      & 0.137&      1.089&      5.439&      0.217&      0.830&      0.942&      0.975 \\
        Monodepth2 \cite{Godard-iccv19}         & 0.115&      0.903&      4.863&      0.193&      \st{0.877}& {0.961}&      0.982 \\
        Zhao et al. \cite{Zhao-cvpr20}          & \st{0.113}& \ft{0.704}&    {4.581}&    {0.184}&      0.871&      {0.961}& \st{0.984} \\
        3DHR \cite{wang20223d}                  & \ft{0.109}& 0.790&      4.656&      0.185&      \ft{0.882}& \st{0.962}&      0.983 \\
        GVO \cite{cao2023learning}              & 0.118&      0.787& \st{4.488}& \st{0.183}&      0.870&      \st{0.962}& \ft{0.985} \\
        \hline
        Monodepth2 + SSC             & 0.115&      0.770&      4.813&      0.188&      0.863&      0.957&      0.983 \\
        GVO + SSC                    & 0.115& \st{0.747}& \ft{4.416}& \ft{0.181}& 0.868& \ft{0.964}&      \ft{0.985} \\
       \hline
    \end{tabular}
\end{table}

\subsection{Ablation Study}
This part presents the ablation analysis to assess the performance of the proposed SIDE method.

\textbf{Experiments on $SSIM_m$.}
We first test the loss function combining with $SSIM_m$ and MAE,
\begin{equation}
    \mathcal{L}_m = (1-w) MAE + w (1-L^\alpha C^\beta (\frac{1}{2}(1+S))^\gamma),
\end{equation}
where $w=0.85$ is similar to previous works \cite{Godard-iccv19,Luo-pami20}.
The baseline is the re-implemented Monodepth2 \cite{Godard-iccv19}, which is trained using $\alpha=1$, $\beta=1$, $\gamma=1$. 
Compared with the original version, the performances of re-implemented method is largely consistent with the original version.

Table \ref{table:SSIM_m} shows the experimental results. We can see that most of the combinations of $\alpha$, $\beta$ and $\gamma$
are better than setting them to 1.
Compared with baseline, most of the indicators are better when the weights are simply revised. 
From Table \ref{table:SSIM_m}, we can summarize the following two conclusions.

\textit{1) The contrast $C$ and structure $S$ in SSIM is important than the luminance $L$.}
From Table \ref{table:SSIM_m}, we can clearly find that
the best performance is obtained only when adjusting (increasing) the contrast and brightness weights.

\textit{2) Increasing all the weights is not necessarily better than increasing a single weight.}
From Table \ref{table:SSIM_m}, we can find that
$\alpha,\beta,\gamma=2$ or $3$ is not better than adding partial weights alone.
One possible reason is that increasing $\alpha,\beta$ and $\gamma$ at the same time
will result in more flat areas, like Figure \ref{fig:examples}(c).
Its areas may make the gradient propagation insignificant and bring convergence difficulties.

\textbf{Experiments on $SSIM_a$.}
We then test the loss function combining with $SSIM_a$ and MAE,
\begin{equation}
    \mathcal{L}_a = w_1 MAE + w_l(1-L) + w_c(1-C) + w_s(1-\frac{1}{2}(1+S)).
\end{equation}

Table \ref{table:SSIM_a} shows the experimental results. 
We can see that most of the combinations of $w_1$, $w_l$, $w_c$ and $w_s$
are better than baseline which uses multiplication rather than addition to fuse SSIM.
From Table \ref{table:SSIM_a}, we summarize the following two conclusions.

\textit{1) SSIM loss constructed by addition outperforms multiplication in terms of upper bound on performance.}
Compared with the best performances of Table \ref{table:SSIM_a} and Table \ref{table:SSIM_m}, 
we can find that most of the best results come from Table \ref{table:SSIM_a}, especially the indicator SqRel.

\textit{2) Statistics reveals that the performances is better when $w_c=0.5$ and $w_s=0.7$.}
We counted the edge distribution of all results in Table \ref{table:SSIM_a} and found that
the performance is better when $w_c=0.5$ and $w_s=0.7$.
The values of $w_1$ and $w_l$ do not significantly affect the results of depth estimation.

\begin{figure}[ht]
    \centering
    \includegraphics[width=10cm]{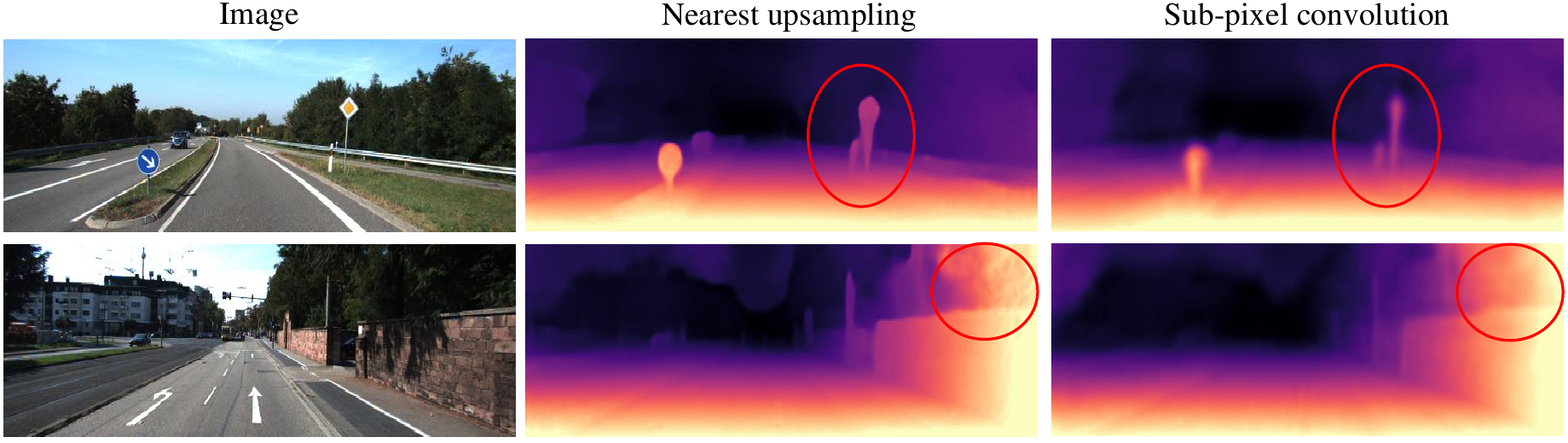}
    \caption{
    Visualization of different upsampling methods.
    }
    \label{fig:vis_upsampling}
 \end{figure}

\begin{figure}[ht]
    \centering
    \includegraphics[width=10cm]{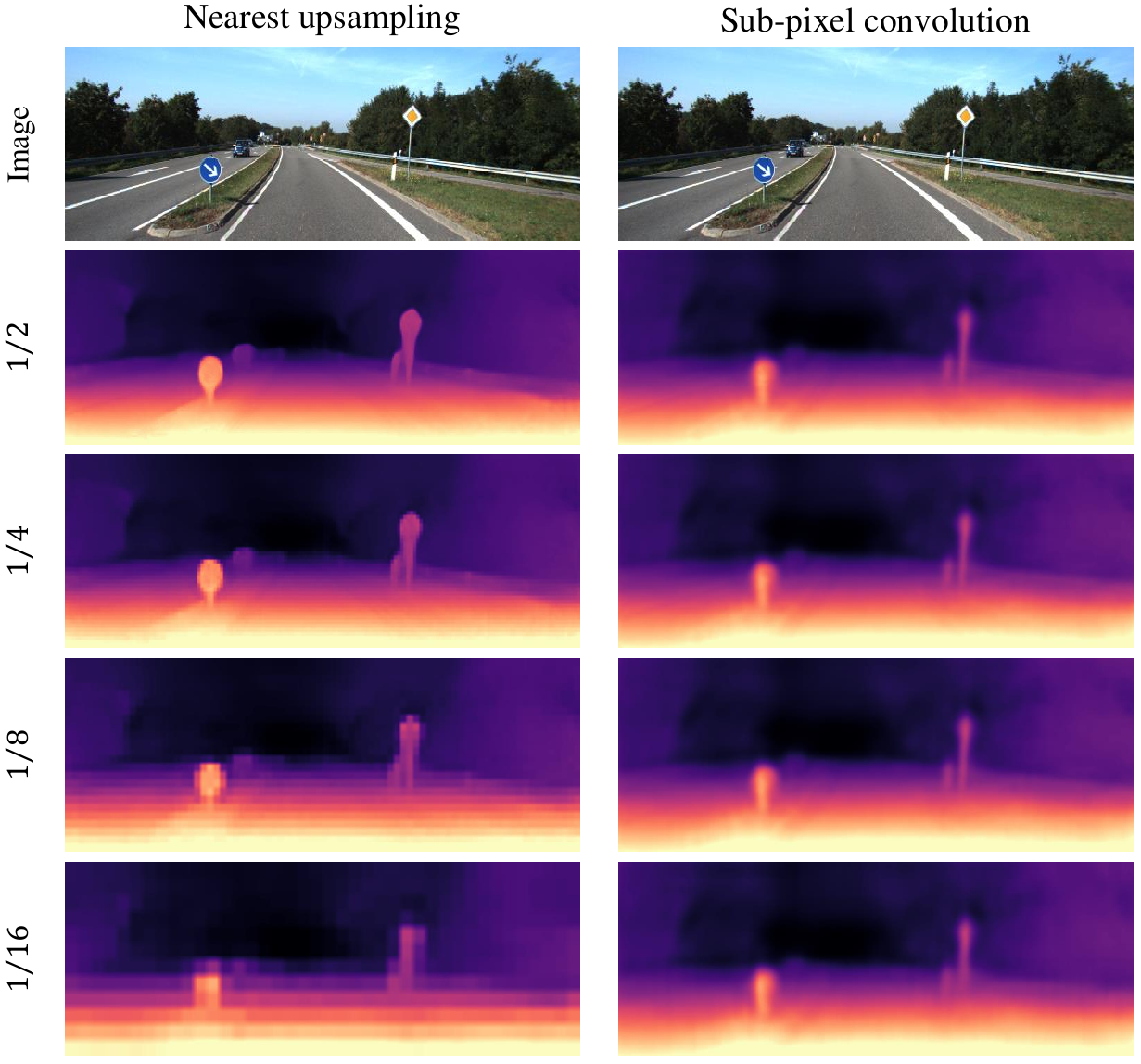}
    \caption{
        Visualization of different upsampling methods at different resolution.
    }
    \label{fig:vis_multi_scale}
 \end{figure}

\textbf{Experiments on sub-pixel convolution}
In this part, we show the experiments on upsampling method.
We substitute nearest upsample with sub-pixel convolution approach.
Table \ref{table:subpixel} and Table \ref{table:resolution} show some experimental results.

\textit{1) Using sub-pixel convolution is better than nearest upsampling on most metrics.}
Table \ref{table:subpixel} shows four groups of experiments. 
We can see that the sub-pixel convolution is better than the nearest upsampling module on most 
metrics, especially on SqRel and RMSlog.
In addition, for the accurate metric $<1.25$, nearest upsampling is always better than sub-pixel convolution.
This may be because the depth estimated by sub-pixel convolution is more smooth and therefore lacks precision,
like the visualization results in Figure \ref{fig:vis_upsampling}.

\textit{2) Sub-pixel convolution still has good performance at small resolutions.}
Table \ref{table:resolution} shows the performances at different resolutions.
As we can see that the sub-pixel convolution has consistent performances at each resolutions.
Since good depth estimates are learned at low resolution, 
it may be more difficult to learn clear contours during constant upsampling decoding.
Figure \ref{fig:vis_multi_scale} shows the visualization results.

\subsection{Comparison with state-of-the-art}
The predicted depths are reported and compared with several methods in Table \ref{table:sota}.
When our modified loss function and upsampling module were incorporated into Monodepth2 \cite{Godard-iccv19} and GVO \cite{cao2023learning} 
(i.e., "Monodepth2 + SSC" and "GVO + SSC" in Table \ref{table:sota}), their performances were largely improved.
We can find that compared with recent methods, 
which use multi-frames \cite{wang20223d} or auxiliary optical flow model \cite{Zhao-cvpr20,cao2023learning},
our model also performs well under some metrics.
Note that the the weights of $SSIM_a$ is set to $w_1=0.4$, $w_l=0.5$, $w_c=0.5$, $w_s=0.7$ in Table \ref{table:sota}.

\section{Discussion}
\textbf{Motivation.}
From the first unsupervised SIDE approach \cite{Zhou-cvpr17} to recent state-of-the-art methods \cite{cao2023learning,Godard-iccv19},
the combination of MAE with SSIM as loss function seems to be a standard option.
However, we found in practice that it is difficult to reflect the change in depth through the measurement of the pixel-wise SSIM loss.
In many cases, different depths will result in similar RGB values by re-projection,
and large changes in depth can only lead to small shifts in SSIM values.
These phenomenons drive us to find a suitable loss function for unsupervised SIDE task.
Modification to the SSIM function is the simplest solution, 
and this work demonstrates the effectiveness of this modification from a large number of experiments.
The future work is to analyze the gaps between the supervised and unsupervised manners, and design a more valid loss function.

\textbf{Generalization.}
Although we only apply the proposed SSIM to Monodepth2 \cite{Godard-iccv19} and GVO \cite{cao2023learning}, 
we think that it can be easily promoted and might improve the performances in many other unsupervised SIDE task.
One reason is that there is no conflict between the proposed loss and the assumptions of the existing models,
which usually develop the occlusion mask \cite{Luo-pami20} and pursuing accurate pose \cite{Zhao-cvpr20,cao2023learning}.

\textbf{Limitation.} 
Although the proposed SSIM is better than original version in every indicator of SIDE evaluation,
the increase of performances is small. The RMS from 4.856 to 4.822 is improved by 0.7\%,
and the SqRel from 0.868 to 0.816 is improved by 6.4\%. 
This shows that there is still a lot of room for futher improvement in the loss function.

\textbf{Sub-pixel convolution.} 
We argue that sub-pixel convolution might decrease the performances in the case of using more powerful model.
We may be aiming for a smooth estimate when the depth is overall inaccurate, 
however, the edge areas, usually the most difficult part to learn, is also the pursuit of future work.
Thus, how to take advantage of sub-pixel convolution and avoid its disadvantages is also a good research direction.

\section{Conclusions}
In this work, we explored the loss function and the upsampling approach for the task of unsupervised SIDE.
Entensive experiments showed that the proposed version of SSIM loss is better than the original one, 
especially the form combining by addition.
Further more, we proposed to use sub-pixel convolution upsampling method instead of nearest interpolation 
for multi-scale training. 
The results showed that the proposed approaches achieve a significant performance improvement compared with the baseline.

\section{Acknowledgements} 
This work was supported by 
National Natural Science Foundation of China (62076055) and
Sichuan Science and Technology Program (2022ZYD0112).

%
%
%
%
\bibliographystyle{splncs04} 
\bibliography{ms}

\end{document}